\documentclass[10pt,twocolumn,letterpaper]{article}

\usepackage{cvpr}
\usepackage{times}
\usepackage{epsfig}
\usepackage{graphicx}
\usepackage{amsmath}
\usepackage{amssymb}
\usepackage{comment}
\usepackage{subfigure}
\newcommand{\chgd}[1]{{\color{red}{#1}}}

\cvprfinalcopy
\makeatletter
\newcommand{\figcaption}[1]{\def\@captype{figure}\caption{#1}}
\newcommand{\tblcaption}[1]{\def\@captype{table}\caption{#1}}
\makeatother
\usepackage[pagebackref=true,breaklinks=true,letterpaper=true,colorlinks,bookmarks=false]{hyperref}

\def\cvprPaperID{494} 

\ifcvprfinal\fi
\begin{document}

\title{Anticipating Traffic Accidents with Adaptive Loss and Large-scale Incident DB}

\author{
\renewcommand{\thefootnote}{\fnsymbol{footnote}}
Hirokatsu Kataoka${}^\text{1}$\thanks {\normalsize {indicates equal contribution}} \hspace{10mm} Tomoyuki Suzuki${}^\text{1,2}$$^*$ 
\hspace{10mm} Yoshimitsu Aoki${}^\text{2}$ \hspace{10mm} Yutaka Satoh${}^\text{1}$\\
 ${}^\text{1}$National Institute of Advanced Industrial Science and Technology (AIST)
\hspace{5mm} ${}^\text{2}$Keio University \hspace{5mm}\\
\tt\small {\{hirokatsu.kataoka, suzuki-tomo, yu.satou\}@aist.go.jp}
\\
\tt\small {tosuzuki@aoki-medialab.jp,} \hspace{1mm} \small{aoki@elec.keio.ac.jp}
}    
\maketitle
\thispagestyle{empty}
\begin{abstract}
   In this paper, we propose a novel approach for traffic accident anticipation through (i) Adaptive Loss for Early Anticipation (AdaLEA) and (ii) a large-scale self-annotated incident database for anticipation. The proposed AdaLEA allows a model to gradually learn an earlier anticipation as training progresses. The loss function adaptively assigns penalty weights depending on how early the model can anticipate a traffic accident at each epoch. 
   Additionally, we construct a Near-miss Incident DataBase for anticipation. This database contains an enormous number of traffic near-miss incident videos and annotations for detail evaluation of two tasks, risk anticipation and risk-factor anticipation.
In our experimental results, we found our proposal achieved the highest scores for risk anticipation (+6.6\% better on mean average precision (mAP) and 2.36 sec earlier than previous work on the average time-to-collision (ATTC)) and risk-factor anticipation  (+4.3\% better on mAP and 0.70 sec earlier than previous work on ATTC).
\end{abstract}

\section{Introduction}
Recently, progress in advanced driver assistance systems (ADASs), including self-driving cars, has been on the rise due to contributions from such fields as computer science, robotics, and traffic science. Among these advanced techniques, advanced computer vision algorithms are especially important for implementation in ADASs. In self-driving cars, the primary objective must be to produce ``a car that carries humans to their destination safely", and one vital technology field for achieving this target, three-dimensional (3D) environment sensing, has seen significant improvements recently. For example, laser sensors such as light detection and ranging (LiDAR) and visual simultaneous localization and mapping (vSLAM) are among the most active topics in the race for practical self-driving cars capable of transporting human passengers. 
\begin{figure}[t]
\centering
   \includegraphics[width=1.0\linewidth]{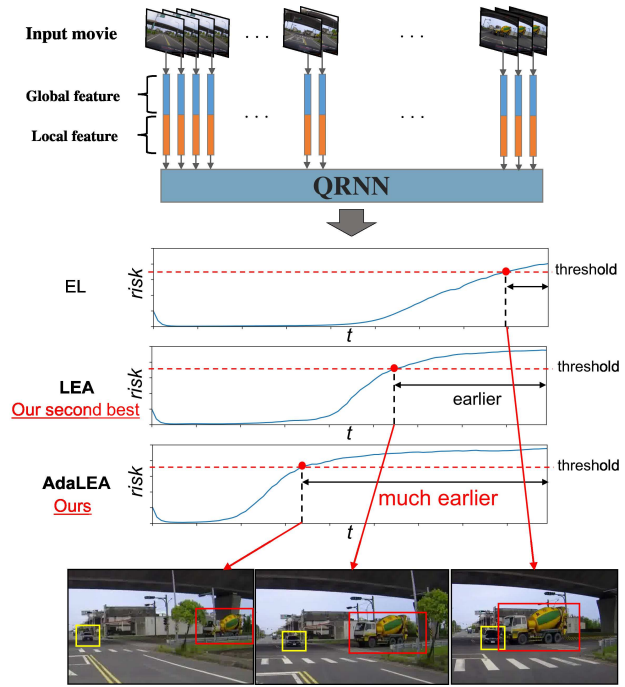}
   \caption{\textbf{Our proposed Adaptive Loss for Early Anticipation (AdaLEA), EL (conventional work), and LEA (also ours).} Our AdaLEA allows a model to anticipate incidents/accidents earlier as training progresses by referring to the ability of its anticipation. We achieved a traffic risk-factor anticipation with 3.65 seconds average time-to-collision (ATTC), versus 2.99 seconds for a conventional EL on NIDB.}
\label{fig:overview}
\end{figure}
Also, two-dimensional (2D) image processing is one of the key to achieve safe driving and has addressed traffic tasks, including pedestrian (object)  detection, semantic segmentation and situational awareness.

For a trail of safe driving, Geiger~\textit{et al.} have collected the KITTI benchmark to evaluate several self-driving car tasks~\cite{geiger2012we}, problems related to semantic segmentation, and two- and three-dimensional (2D/3D) object detection. However, conventional databases represented by this lack traffic accident or near-miss incident (accident/incident) cases, even though the target of a traffic safety system is to avoid dangers. The database of accident/incident videos is necessary to highly understand a traffic danger situation, 
therefore we constructed a novel database that contains a large-number of near-miss traffic incidents with detailed annotations for anticipation.

In addition to database construction, this study explores how to anticipate traffic accident/incident cases. We contend that the key to avoiding accidents/incidents is earlier anticipation in the framework. 
Herein, we propose a traffic accident anticipation model (see Figure~\ref{fig:overview}) that operates through an {\it adaptive} penalty weighted value for early anticipation, in contrast to conventional anticipation learning procedures with a {\it static} one.
As the result of our contributions, we found that our approach achieved risk-factor anticipation with 62.1\% mean average precision (mAP) and 3.65 sec average time to collision (ATTC), which is +4.3\% more accurate and  0.73 sec earlier than conventional work. Note that we define risk-factor as a object that cause a accident (e.g., cyclist, pedestrian and vehicle).

In summary, our contributions are as follows:

\textbf{\underline{Technical contribution}:} We propose our Adaptive Loss for Early Anticipation (AdaLEA) method, which allows a model to gradually learn an earlier anticipation as training progresses, inspired by Curriculum Learning\cite{bengio2009curriculum}. By referring to the ATTC during each training epoch, penalty weight adaptively changes. Moreover, in our base model, we assign a quasi-recurrent neural network (QRNN)~\cite{qrnn} that enables stable output from temporal convolution on consecutive sequential data such as videos, by replacing the long-short term memory (LSTM)~\cite{luo2016efficient} used conventionally.

\textbf{\underline{Database contribution}:} We have annotated a novel traffic Near-miss Incident DataBase (NIDB) that contains a large-number of near-miss traffic incidents to (i) raise awareness of the problem of risk-factor anticipation, and (ii) improve feature representation in anticipation.



\section{Related works}
Since this paper addresses topics such as self-driving cars and temporal anticipation, it is relevant to a large number of areas. However, we limited our focus to closely related and representative topics that are relevant to our work:

\textbf{Anticipation in videos}: Anticipation in videos is a very challenging issue in the field of computer vision because a future event must be anticipated from information up to the present and it's often ambiguous. 
 There is little promising work in video-based risk anticipation. Therefore, herein we touch on work related to our traffic accident anticipation method such as early event detection and anticipation. 

Early event detection is the task that a model should detect the event before it is completed.
The representative work on this topic was conducted by Ryoo~\cite{ryoo2011human}, who introduced a probabilistic model for early event detection. In the context of traffic situation, Kataoka~\textit{et al.} defined transitional action~\cite{KataokaBMVC2016} and constructed pedestrian-specified database~\cite{KataokaSensors2018} for short-term action prediction.
Then, Aliakbarian~\textit{et al.} proposed an early event detection method that uses a spatial attention mechanism~\cite{aliakbarian2017encouraging}.  
Event anticipation is the problem to anticipate the event before it occurs. For example, Koppula~\textit{et al.}~\cite{koppula2016anticipating} propose the anticipation method using CRF with information of human poses and object coordinates, and Vondrick~\textit{et al.}\cite{vondrick2016anticipating} train CNN to extract feature for action anticipation in self-supervised manner. Accident anticipation belongs to event anticipation, since accident/incident must be anticipated before  occurrence to avoid them. In the area of risk anticipation, a number of other sophisticated algorithms have recently been proposed. For example, Chan~\textit{et al.} introduced the concept of  dynamic soft-attention (DSA) involving an LSTM to anticipate traffic accidents~\cite{ChanACCV2016}, and Zeng~\textit{et al.} have improved target-focused risk anticipation by introducing Imaging future mechanism, which predict future location of the target~\cite{ZengCVPR2017}. In these two works of risk anticipation, Exponential Loss (EL) proposed by Jain~\textit{et al.}~\cite{jain2016recurrent}, which changes the penalty weight in accordance with the difficulty at each frame, is utilized for training a model. However this loss function does not encourage {\it earlier} anticipation since this always gives higher weights only frames close to the accident.
 

Since our philosophy aims at avoiding events in advance, we must execute an anticipation as {\it early} as possible. In this paper, we try to accomplish this via our new loss function AdaLEA. Moreover, we also employ QRNN~\cite{qrnn} in lieu of more frequently used LSTM~\cite{luo2016efficient}. In the experimental section, we show the effectiveness of them, in comparison with above mentioned algorithms~\cite{jain2016recurrent,ChanACCV2016,ZengCVPR2017}.

\begin{figure*}
	\begin{center}
	\includegraphics[width=0.95\linewidth]{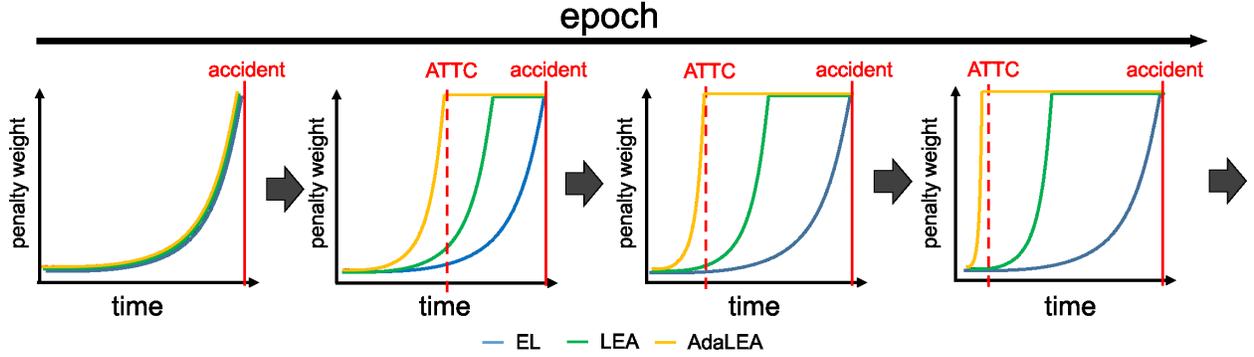}
	\end{center}
   	\vspace{-10.0pt}\caption{At first training epoch, EL (blue), LEA (green) and AdaLEA (yellow) assign equal penalty weights and weights of EL are static at all training epochs. According to training progress, a penalty weight of our second-best LEA are linearly shifted to promote early anticipation and our AdaLEA flexibly changes penalty weights depending on the validated ATTC (Average Trame-to-Collision) at each previous epoch.}
	\label{fig:overview_of_loss}
\end{figure*}


\textbf{Traffic database}: Several practical databases for traffic safety have been proposed in the past decade. In the pedestrian database, Dollar \textit{et al.} released a large-scale and realistic Caltech pedestrian dataset~\cite{DollarCVPR2009,DollarPAMI2012} that has proved to be beneficial for improving the local descriptors, classifiers, and models. Note that detailed analysis, such as occlusion rates, data statistics, and burden comparisons, are areas of extensive study in the pedestrian detection field.
In 2012, the KITTI benchmark was applied to set meaningful vision problems for self-driving cars~\cite{geiger2012we}. These include problems in optical flow, semantic labeling, visual odometry, stereo vision, 2D/3D object detection, and temporal tracking. Thanks to the sophisticated approaches now available, such as fully convolutional networks (FCN)~\cite{LongCVPR2015} and region-based convolutional neural networks (R-CNN)~\cite{GirshickCVPR2014}, there has been improved performance of solving these problems using the KITTI benchmark. 
Another interesting recent impact is the Toronto City dataset~\cite{WangICCV2017}, which uses a huge amount of data obtained via various sensors for large-scale city reconstruction. The use of different sensor types provides a variety of perspectives that can be applied to comprehensive auto navigation matters.
However, these representative databases contain few scenes that present accidents/incidents in which pedestrians, cyclists, or other vehicles must be avoided before mishaps occur. 

Dashcam Accident Dataset (DAD)~\cite{ChanACCV2016} contains accidental events on the collected data. However, it is not large enough (there are only $10^{2}$-order accident videos, which are 5 seconds each) to optimize a high-level model, 
and many of accidents in DAD are caused between others (e.g., other motorcycle-other vehicle), not own vehicle and other that should be avoided by own vehicle.
Thus, there is an urgent need for a collection of large-scale incident scenes with annotations to ensure that a self-driving car can learn how to safely navigate dangerous situations. 
In this paper, we constructed a novel database that contains a large-number of near-miss traffic incidents with detailed annotations, especially for risk anticipation task that is the one of the key problem to the implementation of self-driving car.

\section{Our Approach}

The overview of our system is shown in Figure~\ref{fig:overview}. 
The system extracts global and local feature from each frame, executes temporal analysis on them and output risk rate at each frame that represents probability that an accident will occur in the future.
To the model we introduce QRNN, which enables a model to achieve stable anticipation with temporal convolution on consecutive features. 
For training the model, we use our novel loss function, Adaptive Loss for Early Anticipation (AdaLEA). 
In this section, we explain AdaLEA, which is the primary contribution of this study, and introduction of QRNN instead of LSTM for temporal analysis in anticipation tasks. The details of global and local feature are in section~\ref{sec:eval}.


\subsection{Loss function}
To avoid danger in advance, traffic accident anticipation needs both accuracy and earliness.
Figure~\ref{fig:overview_of_loss} shows the overview of three different loss functions. Our strategy is to adaptively modify the weight value depending on how early the model can anticipate a traffic accident at each learning epoch. The flexible operation of our AdaLEA allows us to provide an earlier anticipation than other approaches with conventional Exponential Loss (EL)~\cite{jain2016recurrent} and our second-best Loss for Early Anticipation (LEA).
	
	In a loss function for anticipation, uniformed weighting is susceptible to unstable learning since the difficulty to anticipate varies over the time. To resolve the problem, we design our losses based on EL~\cite{jain2016recurrent}. The training module with EL changes the penalty weight in accordance with the difficulty at each frame in order to stabilize an anticipation learning.
	However EL does not encourage early anticipation at all since the function always gives a higher weights only close to the accident (see blue in Figure~\ref{fig:overview_of_loss}), therefore we introduce a mechanism for early anticipation.
	The losses we propose are divided into positive (a video including a traffic accident) and negative (no accident, a normal driving scene) samples.
	While the loss for the negative sample is standard cross-entropy, a weighting value in the positive sample is gradually increased when a video frame is closer to an accident frame like EL. 
	Moreover, we utilize the idea of Curriculum Learning~\cite{bengio2009curriculum} which ranges from easy to difficult samples in training time and improves generalization of a model. In anticipation, an easy sample is a frame close to an accident time (e.g., a few frames from an annotated accident/incident time) and a difficult sample is one farther away (e.g., over 5 seconds from an accident time), that is to say early anticipation.  
	According to this, for smooth optimization, our losses allow a model to gradually anticipate earlier as training progress.
	The LEA is shown as below:

	\mbox{\underline {\textbf {\textit{Loss for Early Anticipation}}}}
	
	\mbox{\it{for positive}} : 
	\begin{eqnarray}
		&\mbox{\rm{L}}_{LEA}^{p}(\{r_t\})=\sum_{t=1}^{T} -\alpha \mbox{\rm{log}}(r_t)\\
		&\alpha = \mbox{\rm{exp}}(-\mbox{\rm{max}}(0, d - \lambda (e-1)))
	\end{eqnarray}

\begin{figure*}[t]
\centering
\subfigure[Penalty weights in EL and LEA]{\includegraphics[width=0.425\linewidth]{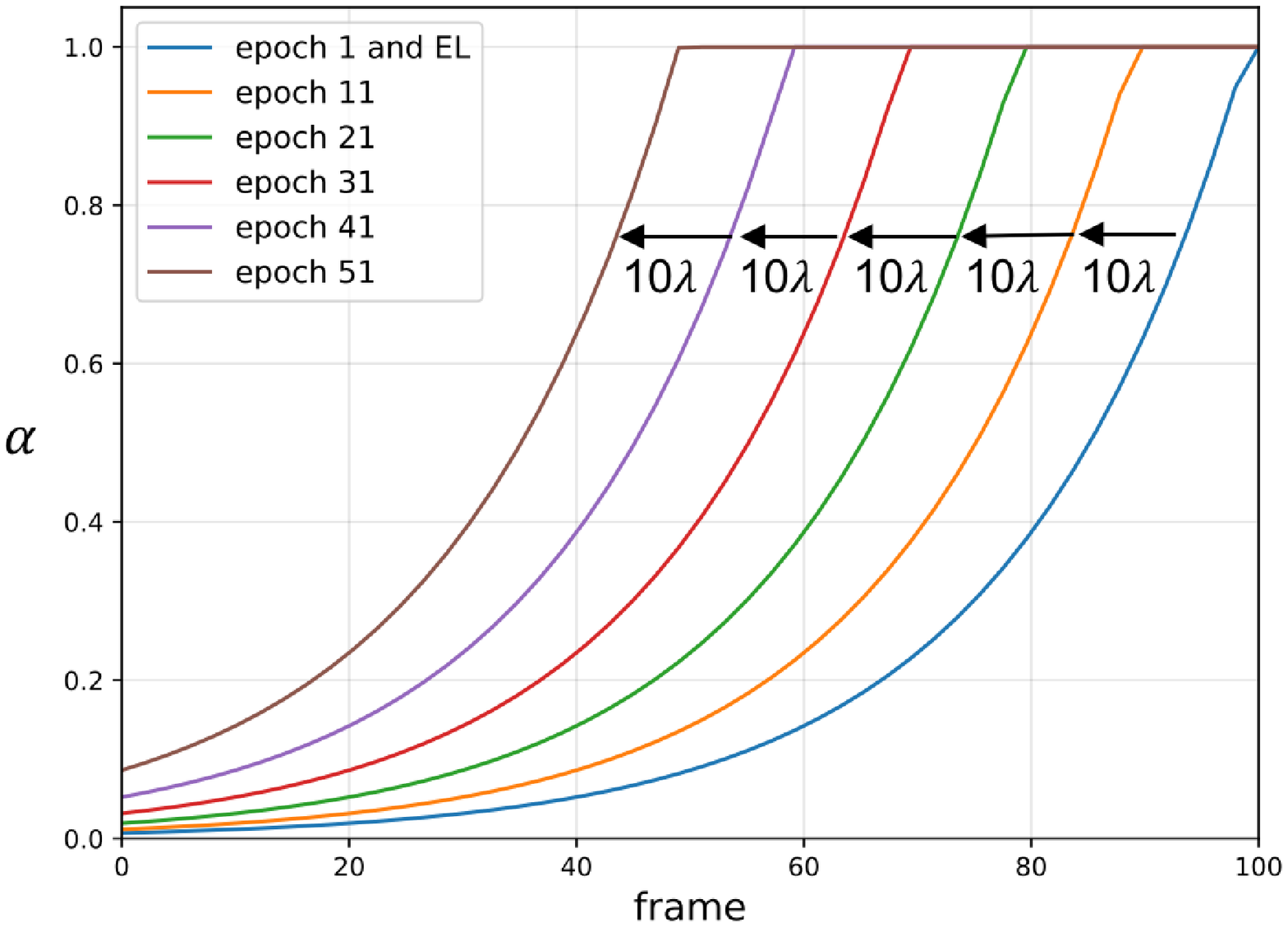}
\label{fig:LEA}}
\subfigure[Penalty weights with our AdaLEA]{\includegraphics[width=0.445\linewidth]{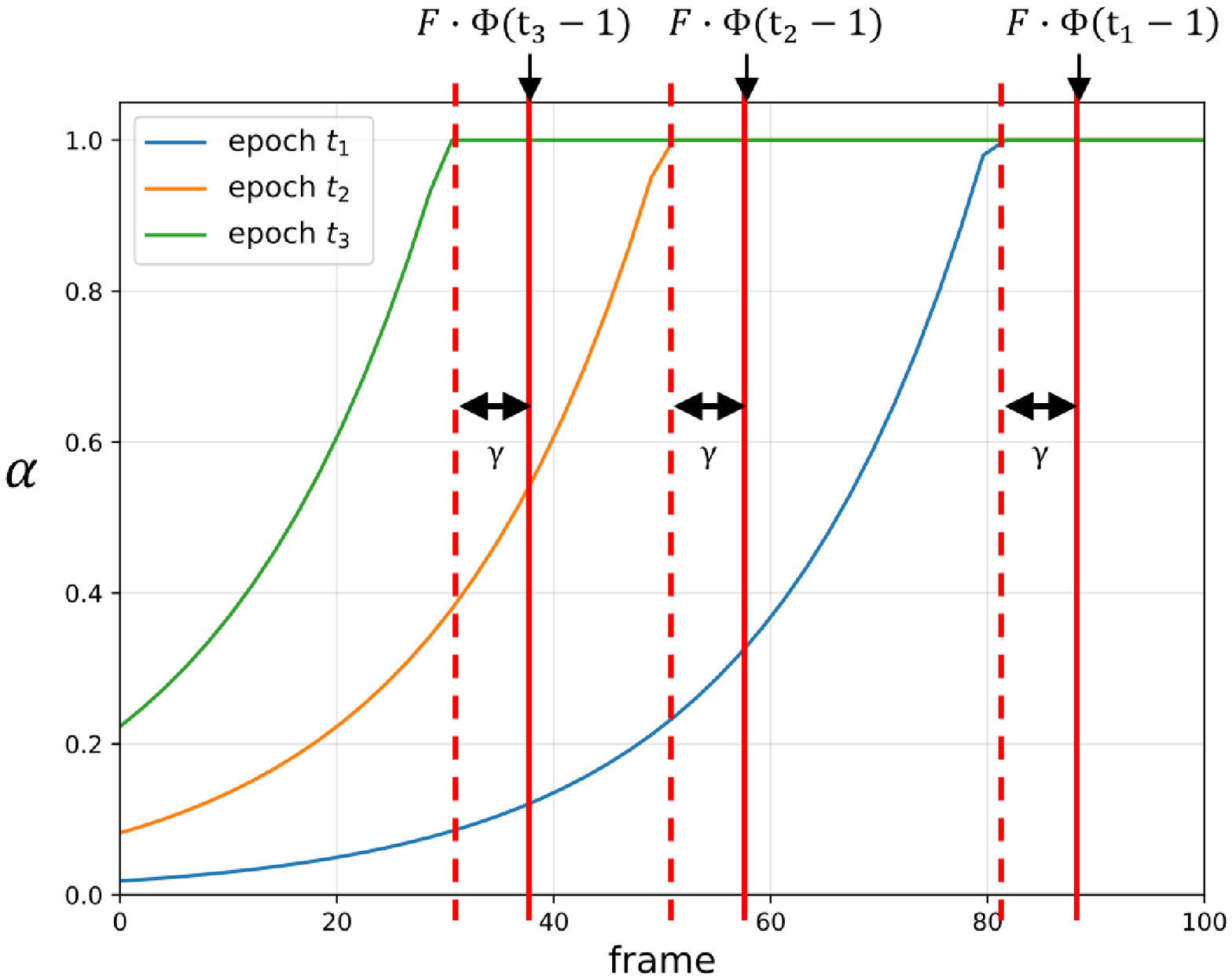}
\label{fig:AdaLEA}}
\caption{Detailed progresses of penalty weights in EL, LEA and AdaLEA in case that videos contain 100 frames including accident/incident at the last: (a) Penalty weights of LEA are linearly shifted depending on epoch and factor of proportionality $\lambda$. Penalty weights of EL(blue line) are static at all training epoch and equal to LEA at the epoch 1. (b) To promote an earlier anticipation, penalty weights of AdaLEA are adaptively changed depending on a validated ATTC (from $\Phi(\cdot)$, solid red lines) at each previous epoch and hyper parameter $\gamma$. Note that dashed red lines indicate the just time when penalty weights saturate to $1$.}
\label{fig:lossfunction}
\end{figure*}

	\mbox{\it{for negative}} :
	\begin{eqnarray}
		&\mbox{\rm{L}}_{LEA}^{n}(\{r_t\})=\sum_{t=1}^{T} -\mbox{\rm{log}}(1-r_t)
	\end{eqnarray}
	where $r_{t}$ indicates risk rate in range $[0, 1]$ at video time $t$, $T$ is starting frame of annotated accident/incident, and $d=T-t$ which means the frames from current frame ($t$) to accident/incident ($T$). Moreover, $e$ represents a current learning epoch, and $\lambda$ is a hyper-parameter. 
	The penalty weights of anticipation at early time $t$ is weak in an early learning stage, but increases according to the learning progress. Figure~\ref{fig:LEA} shows LEA in a different learning epoch. Note that at the beginning of training ($e = 1$) or if $\lambda = 0$, the LEA is equal to the EL~\cite{jain2016recurrent}.

LEA is designed based on the premise that according to training progress the model can anticipate an accident earlier {\it linearly},
however indeed this premise is not always right due to various learning situation. Therefore, we further propose AdaLEA that provides adaptive penalty value, depending on the anticipation time,  by referring to ATTC (see section~\ref{subsec:setting}), that represents how early a model anticipates in average. AdaLEA is given as below:
	

    \mbox{\underline {\textbf {\textit{Adaptive Loss for Early Anticipation}}}}

  	

	%
	\mbox{\it{for positive}} : 
	\begin{eqnarray}
		&\mbox{\rm{L}}_{AdaLEA}^{p}(\{r_t\})=\sum_{t=1}^{T} -\alpha \mbox{\rm{log}}(r_t)\\
		&\alpha = \mbox{\rm{exp}}(-\mbox{\rm{max}}(0, d - F\cdot{}\Phi(e-1) - \gamma))
	\end{eqnarray}

	\mbox{\it{for negative}} :
	\begin{eqnarray}
		&\mbox{\rm{L}}_{AdaLEA}^{n}(\{r_t\})=\sum_{t=1}^{T} -\mbox{\rm{log}}(1-r_t)
	\end{eqnarray}
	where $\Phi(\cdot)$ is a function which represents an ATTC at a training epoch, $F$ is the frame rate of videos, and $\gamma$ is a hyper-parameter. In short, the loss is adaptively penalized depending on the ability of early anticipation in order to promote the training process. Figure~\ref{fig:AdaLEA} shows examples of penalty weights at three learning epochs. Here, it can be that the AdaLEA makes an anticipating system earlier than in the previous epoch at all training times.

\subsection{Quasi-recurrent neural networks (QRNN)}
To analyze continuous sequential data like videos, motion-information that can be obtained considering the relationship between adjacent times is important. Although LSTM is still used for standard temporal analysis in risk anticipation~\cite{ChanACCV2016,ZengCVPR2017,aliakbarian2017encouraging}, they cannot always account for {\it direct} relationships between adjacent frames. Therefore, in lieu of LSTM, we selected QRNN~\cite{qrnn} that includes temporal convolution in order to identify motion-information from direct relationship between adjacent frames.

On natural language processing tasks such as sentiment classification and machine translation, QRNN keeps level of accuracy comparable to LSTM with significant improvement of computational speed. On the other hand, to the best of our knowledge, there is no method that applies QRNN to tasks using consecutive sequential data such as videos. In this case, as mentioned in the beginning of this section, we can expect QRNN to provide not only faster processing but better accuracy than LSTM because of its temporal convolution. Finally, our system outputs risk rate $r_t$ in range $[0,1]$ though a fully connected layer followed by a sigmoid function at every frame. 
\section{Near-miss Incident DataBase (NIDB) for anticipation}
\label{sec:database}

We have constructed NIDB for traffic accident anticipation based on the original traffic database~\cite{KataokaICRA2018}. We have annotated near-miss incident duration and the bounding boxes of risk-factors in addition to the original traffic database especially for the detail evaluation of anticipation.
Overall, the original database contains over 6.2K videos and 1.3M frames, many of which show incident scenes. The videos were captured using vehicle-mounted driving recorders. The videos are divided into four classes, including \{cyclists, pedestrians, vehicles\} as well as a background (negative) class. Moreover, the near-miss incident duration and bounding boxes of risk-factors are annotated in the large-scale video database. After these annotations are terminated, all elements including near-miss incidents, their durations, and bounding boxes are validated by extra annotators. The detailed database construction is described below: 

\begin{figure}[t]
\begin{center}
\includegraphics[width=1.0\linewidth]{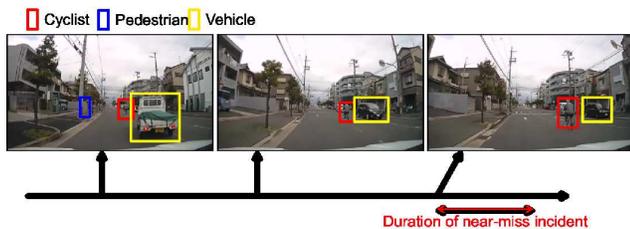}
\end{center}
   \vspace{-15.0pt}\caption{Video annotation with bounding boxes and duration of traffic near-miss incidents.}
\label{fig:videoannotation}
\end{figure}

\subsection{\textbf{Two tasks for traffic accident anticipation}}

	\textbf{Traffic risk anticipation}: This task is following by conventional studies such as Chan~\textit{et al.}~\cite{ChanACCV2016}. Given a video, an anticipation system outputs the probability of a future accident (risk rate) $r_{t}$ in range $[0,1]$ at each frame $t$. We decide the presence of a future accident based on whether $r_{t}$ exceeds the defined threshold $q$ at any frame until the last. Additionally, we define the time-to-collision (TTC) as the period between the time when $r_{t}$ exceeds $q$ and when an accident occurs. The goal of this task is to make a correct anticipation of the potential of an accident occurrence in as long a TTC as possible.

	\textbf{Traffic risk-factor anticipation}: 
	Our NIDB for anticipation provides an additional task, traffic risk anticipation for each risk-factor. 
	In this task, an anticipation system must anticipate what will cause the accident (i.e., what should be payed attention to) in addition to the presence of a accident.
	We use the same procedure for each risk-factor in order to evaluate the correctness of anticipation and TTC. If a model is trained on this task, it can anticipate multiple accidents caused by different risk-factors or no accidents, therefore the model outputs risk rate in range $[0,1]$ for each factor.
The goal of this task is to produce correct anticipations of the potential accidents caused by each risk factor  with the longest TTCs possible.

\subsection{Video annotation for anticipation}

We added two more important annotations to be used when executing a traffic accident anticipation, namely near-miss incident duration and the bounding boxes of candidates of risk-factors (Figure~\ref{fig:videoannotation}). The duration is annotated based on the traffic near-miss incident definition. Annotators place starting and ending times on each video that indicate when they consider a near-miss incident to have occurred and finished, respectively. The bounding boxes are first processed by an accurate object detection algorithm~\cite{ren2015faster} trained on Pascal VOC 2007 dataset~\cite{everingham2015pascal} and we select and modify  bounding boxes and their categories of detected objects. Note that the categories are limited to \{cyclists, pedestrians, vehicles\}. In other elements such as video collection and cross-check, we followed the original traffic database.

Finally, we had collected 4,594 near-miss incidents and 1,650 background videos consisting 100 frames including accident/incident at the last. In the experiment, we randomly split this database into training and testing, where 4,995 training videos (3,675 positive and 1,320 negative) and 1,249 testing clips (919 positive and 330 negative).

\section{Evaluation}
\label{sec:eval}

In this section, we evaluate our proposals on a conventional database~\cite{ChanACCV2016} and our NIDB for anticipaition.

\subsection{Settings\label{subsec:setting}}


\textbf{Database.} We use two databases for traffic accident anticipation.

Dashcam Accident Dataset (DAD)~\cite{ChanACCV2016} contains diverse accidents captured across six cities in Taiwan with dashcam mounted on vehicles. The database consists of 596 positive videos that include accident scenes covering the last 10 frames and 1,137 negative videos without accidents. 
Videos in the database are separated into 1,266 training videos (446 positives, 820 negatives) and 467 testing videos (150 positives and 317 negatives). 

NIDB is our proposed database. The detailed properties are discussed in section~\ref{sec:database}. Moreover, for extraction of global feature we introduce a pre-trained model on our NIDB in addition to the ImageNet~\cite{ImageNet}/Places365~\cite{Places365} pre-trained models in section~\ref{sec:exploration}. We use the NIDB to evaluate the traffic risk anticipation, and risk-factor anticipation.

\textbf{Implementation details and evaluation metrics.} We use deep activation features (DeCAF)~\cite{donahue2014decaf} from VGGNet~\cite{simonyan2014very} for both local  and global features in a traffic scene. 
In the global feature, we directly extract a DeCAF from a full-image. For feature extractor, we employ ImageNet pretrained model and our NIDB-pretrain described in the next subsection.
In the local feature, we use a conventional dynamic soft-attention (DSA)~\cite{ChanACCV2016}, which is the object-specified attention mechanism, in addition to the DeCAF from regions of objects in a traffic scene using ImageNet pretrained model. Finally, we concatenate these two features and input to QRNN or LSTM.

We detect risk-factors such as cyclists, pedestrians, and vehicles by using Faster R-CNN~\cite{ren2015faster}. We train the model on Pascal VOC 2007 dataset~\cite{everingham2015pascal} and fine-tune the detector on each database. 
For the other two hyper-parameters,  we experimentally set $\lambda=3$, $\gamma=5$.

In accident anticipation, both accuracy and earliness are required. We employ the mAP and ATTC by following the previous work~\cite{ChanACCV2016}. 
For each threshold $q$, we can compute precision, recall and TTC. Note that we can compute TTC only for the true positives. By changing the threshold $q$, we can collect many triplets of them and plot the precision v.s. recall and TTC v.s. recall curves. Given these curves, by taking average across different recall, we can compute the mAP and ATTC.

\begin{table}[t]
\begin{center}
\begin{tabular}{rlcccc}
\hline
Units & Pretrain & Acc. & Prec. & Rec. & F-score \\
\hline\hline
 4096 & IN+NIDB & 50.45 & 66.37 & 50.55 & 51.58 \\
 4096 & P+NIDB & 56.00 & 66.71 & 56.00 & 56.82 \\
\hline
64 & P+NIDB & 45.54 & 58.56 & 45.55 & 46.54 \\
128 & P+NIDB & 51.81 & 63.30 & 51.82 & 53.44 \\
256 & P+NIDB & 54.36 & 64.79 & 54.36 & 55.62 \\
512 & P+NIDB & 55.90 & 67.20 & 55.91 & 56.90 \\
1024 & P+NIDB & \textbf{58.54} & \textbf{68.65} & \textbf{58.55} & \textbf{59.86} \\
2048 & P+NIDB & 56.36 & 67.59 & 56.36 & 57.02 \\
4096 & P+NIDB & 56.00 & 66.71 & 56.00 & 56.82 \\
\hline
\end{tabular}
\end{center}
\caption{Comparison of representative pre-trained models and additional NIDB training (IN: ImageNet, P: Places365)}
\label{tab:nidbpretrain}
\end{table}

\begin{table*}[h]
  \begin{minipage}[t]{.45\textwidth}
    \begin{center}
      \scalebox{0.75}{
		\begin{tabular}{ccccc|ccc}
		&Chan16&Zeng17&Chan16&QRNN&Ours1&Ours2&Ours3\\
		& ~\cite{ChanACCV2016} & ~\cite{ZengCVPR2017} &+AdaLEA&+EL&LEA&AdaLEA&+NIDB\\
		\hline\hline
		DSA & \checkmark & -- & \checkmark & \checkmark & \checkmark & \checkmark & \checkmark \\
		LSTM & \checkmark & \checkmark & \checkmark &  &  &  &  \\
		QRNN &  &  &  & \checkmark & \checkmark & \checkmark & \checkmark \\
		EL & \checkmark & \checkmark &  & \checkmark &  &  &  \\
		\textbf{LEA} & &  & \checkmark &  & \checkmark &  &  \\
		\textbf{AdaLEA} &  & &  &  &  & \checkmark & \checkmark \\
		\textbf{NIDB} &  & &  &  &  & & \checkmark \\
		\hline\hline
		mAP[\%] & 48.1 & 51.4 & 49.2 & 51.7 & 52.1 & 52.3 & \textbf{53.2} \\
		ATTC[s] & 1.34 & 3.01 & 2.80 & 3.02 & 3.22 & 3.43 & \textbf{3.44} \\
		\end{tabular}
		}
	    \end{center}
	    \caption{Results of risk anticipation on DAD: NIDB in the table means NIDB-pretrain for global feature. The result of conventional methods are cited from ~\cite{ZengCVPR2017}.}
	    \label{table1}
  \end{minipage}
  \hfill
  \begin{minipage}[t]{.45\textwidth}
    \begin{center}
      \scalebox{0.75}{
		\begin{tabular}{cccc|ccc}
		&Chan16&Chan16&QRNN&Ours1&Ours2&Ours3\\
		&\cite{ChanACCV2016}&+AdaLEA&+EL&LEA&AdaLEA&+NIDB\\
		\hline\hline
		DSA & \checkmark & \checkmark & \checkmark & \checkmark & \checkmark & \checkmark \\
		LSTM & \checkmark  & \checkmark &  &  &  &  \\
		QRNN &  &  & \checkmark & \checkmark & \checkmark & \checkmark \\
		EL & \checkmark &   & \checkmark &  &  &  \\
		\textbf{LEA} & & \checkmark &  & \checkmark &  &  \\
		\textbf{Ada-LEA} &  &  &  &  & \checkmark & \checkmark \\
		\textbf{NIDB} &  &  &  &  & & \checkmark \\
		\hline\hline
		mAP[\%] & 92.5 & 94.4 & 94.2 & 96.2 & 96.3 & \textbf{99.1}  \\
		ATTC[s] & 2.45 & 4.62 & 2.85 & 4.67 & 4.72 & \textbf{4.81}  \\
		\end{tabular}
		}
		\end{center}
		\caption{Results of risk anticipation on NIDB: NIDB in the table means NIDB-pretrain for global feature.}
		\label{table2}
  \end{minipage}
  
\end{table*}

\begin{figure*}[t]
\begin{tabular}{ccc}

		 \begin{minipage}{0.32\hsize}
			\includegraphics[width=0.97\linewidth]{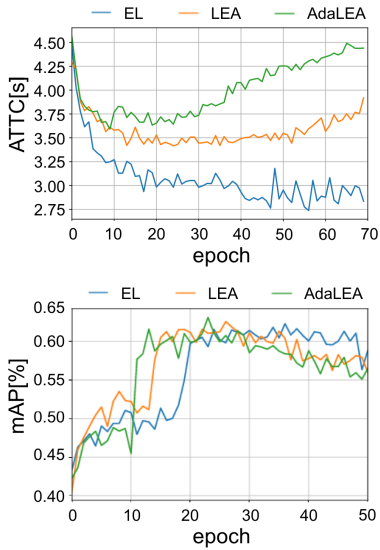}
			\label{fig:mAPvsEPOCH}
			\figcaption{ATTC v.s. epoch (upper) and mAP v.s. epoch (lower) curves.}
		 \end{minipage}

	\def\@captype{table}
	\begin{minipage}{0.68\textwidth}
		\begin{tabular}{c|c|c|ccccc}
		\multicolumn{2}{c|}{}&Chan16~\cite{ChanACCV2016}&Chan16&QRNN&Ours1&Ours2\\
		\multicolumn{2}{c|}{}& &+AdaLEA&+EL&LEA&AdaLEA\\
		\hline\hline
		\multicolumn{2}{c|}{DSA} & \checkmark & \checkmark & \checkmark & \checkmark & \checkmark  \\
		\multicolumn{2}{c|}{LSTM} & \checkmark  & \checkmark &  &  &   \\
		\multicolumn{2}{c|}{QRNN} &  &  & \checkmark & \checkmark & \checkmark  \\
		\multicolumn{2}{c|}{EL} & \checkmark &   & \checkmark &  &    \\
		\multicolumn{2}{c|}{\textbf{LEA}} & &  &  & \checkmark &   \\
		\multicolumn{2}{c|}{\textbf{Ada-LEA}} & & \checkmark &  &  & \checkmark \\
		\hline\hline
		mAP[\%] & bicycle & 57.3 & 57.7 & 60.0 & \textbf{62.8} & 56.8  \\
		    & pedestrian & 43.1 & 43.9 & 43.6 & 44.7 & \textbf{47.9}  \\
		        & vehicle & 73.2 & 75.9 & 79.9 & 78.5 & \textbf{81.4}  \\
		        & Average & 57.8 & 59.2 & 61.2 & 62.0 & \textbf{62.1}   \\
		
		\hline
		ATTC[s] & bicycle & 2.94 & 3.38 & 3.51 & 3.22 & \textbf{3.65}   \\
		     & pedestrian & 3.15 & 3.36 & 3.34 & 3.23 & \textbf{3.56}   \\
		        & vehicle & 2.71 & 2.96 & 2.12 & \textbf{3.99} & 3.75   \\
		        & Average & 2.95 & 3.23 & 2.99 & 3.48 & \textbf{3.65}   \\

		\end{tabular}
		\tblcaption{Results of risk-factor anticipation on NIDB: we used NIDB-pretrain model to extract global feature in all methods.}
		\label{table3}
	\end{minipage}

\end{tabular}
\end{figure*}

\begin{figure*}[t]
\begin{center}
\includegraphics[width=0.95\linewidth]{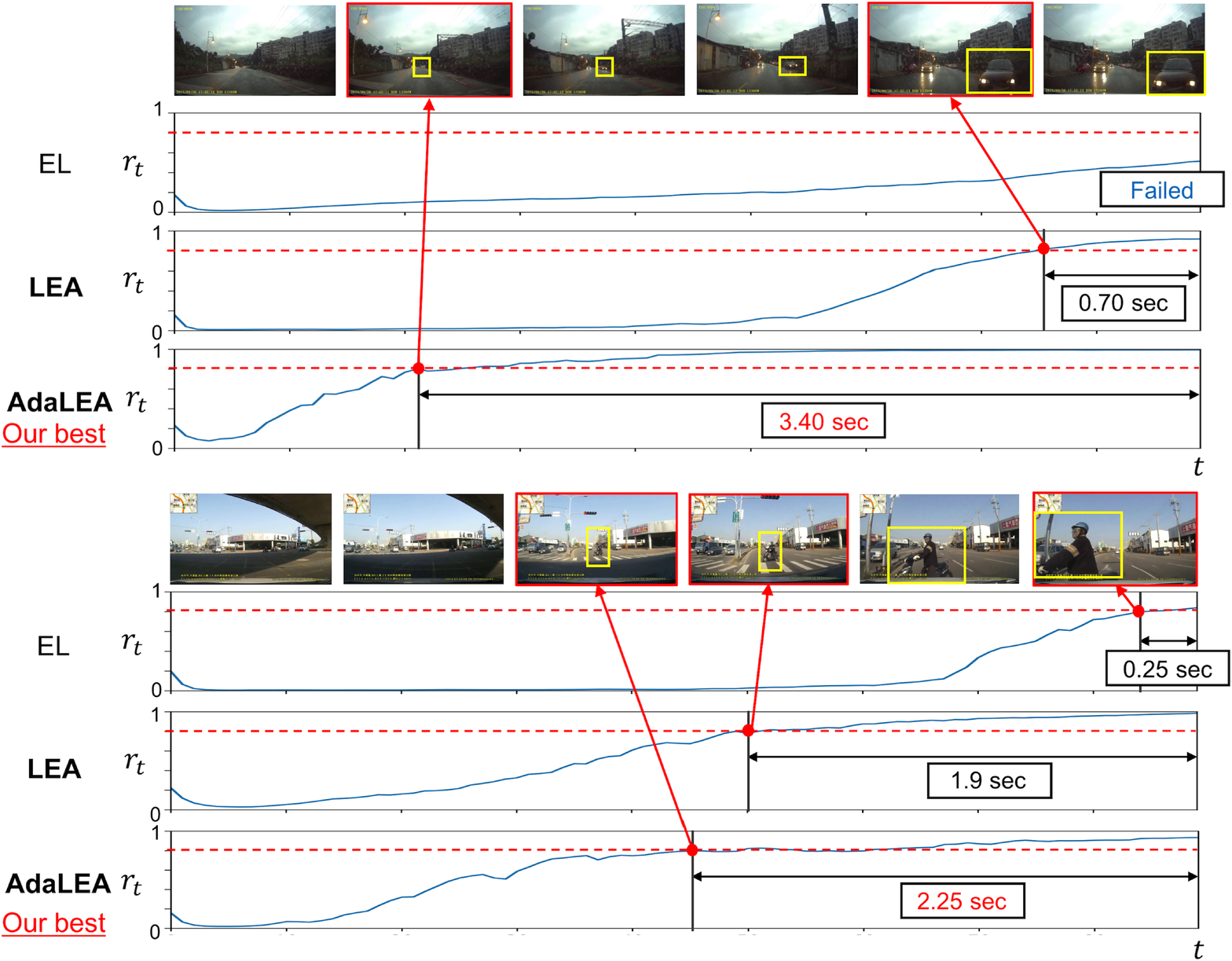}
\end{center}
   \vspace{-15.0pt}\caption{Visual comparison among EL, LEA (our second-best) and AdaLEA (ours) on DAD: Each of image sequences and three bottom graphs shows the example of traffic risk anticipation when we set 0.8 as the threshold. Yellow bounding boxes indicate risk-factors of each video.
   }
\label{fig:vis}
\end{figure*}

\subsection{Exploration}
\label{sec:exploration}

\textbf{Exploration of AdaLEA.} Figure \chgd{5} compare our AdaLEA with conventional EL and our second-best LEA on risk-factor anticipation task of NIDB. To ensure a fair evaluation, all parameters and models used were the same and only the loss function was changed. 
Although the performance rate with mAP was comparable in EL, LEA and AdaLEA (see the lower of Figure \chgd{5}), our AdaLEA achieved earlier traffic accident anticipation than other loss functions (see the upper of Figure \chgd{5}). The differences between our LEA/AdaLEA and EL were linked to gradually changing weighting values at each epoch. Note that in early stage of training, ATTC is high while the mAP is low, and this means that a right evaluation of ATTC is possible after getting a certain mAP (e.g., after around 20 epochs in Figure \chgd{5}).
Ultimately, we determined that the AdaLEA is the most advanced approach since the function gives an adaptive penalty depending on the ability of early anticipation.

\textbf{Exploration of NIDB-pretrain.} Next, we compared the representative pre-trained models (ImageNet and Places365) with 
NIDB-pretrain using various fully connected units to find the optimal model for global feature extractor. 
NIDB-pretrain is the global feature extractor pretrained on per-frame risk-factors \{cyclist, pedestrian, vehicle\} and background classification task (not anticipation), in addition to classification task on Places or ImageNet for traffic scene-specilized feature.
For comparison, we extract global feature per-frame from the last layer of each NIDB-pretrain with various number of units and train SVM to classify risk-factors and background on test split of NIDB. Note that for training NIDB-pretrain, we use only train split of NIDB.
Table~\ref{tab:nidbpretrain} shows the result and we find that 
 an Places365+NIDB extracts the best global feature for near-miss incident classification. Finally, we find that the 1,024-dim setting performed the best rate on the NIDB. Hereafter we will employ Places365+NIDB with 1,024-dim as NIDB-pretrain, and conventional pretrain model on ImageNet~\cite{ImageNet} with 4,096-dim as else standard for global feature extractor. Note that we use conventional pretrain feature on ImageNet for local feature.


\subsection{Comparison with state-of-the-art approaches}

Here, we simply enumerate various base models (DSA, LSTM, QRNN), loss functions (EL, LEA, AdaLEA), and representation  with pre-trained database (NIDB-pretrain). In the base models, we employed the above-mentioned DSA and standard LSTM. Here, the agent-centric risk assessment (ACRA; Zeng17 in Table~\ref{table1}) is used to predict the future coordinates of a target~\cite{ZengCVPR2017}, in addition to the EL and LSTM. 
Note that this method can be applied to the situation that a target which can suffer the danger, namely 'agent' is designated and showed up in video frames, therefore we do not compare this method on NIDB, where 'agent' is always the own vehicle and not showed up.
One more conventional work is Chan16~\cite{ChanACCV2016} which is constructed by \{DSA, LSTM, EL\}. With these methods, we could update our anticipation model and compare it with state-of-the-art approaches~\cite{ChanACCV2016,ZengCVPR2017}, simultaneously.

The quantitative results of traffic accident anticipation on the DAD and NIDB are shown in Table~\ref{table1} and Table~\ref{table2}, respectively, and risk-factor anticipation is shown in Table~\ref{table3}.
In conclusion, we found that our proposed configuration \{DSA, QRNN, AdaLEA, NIDB-pretrain\} achieved the best performance in terms of mAP (53.2@DAD, 99.1@NIDB) and ATTC (3.44@DAD, 4.81@NIDB). In a comparison with the (best) conventional work, the results show our proposal is +1.8@DAD, +6.6@NIDB better with mAP, and +0.43@DAD, +2.36@NIDB earlier with ATTC. In the ATTC on both databases, we can see an especially remarkable value. Since the DAD contains the 4.5-second videos, the value 3.01 seconds with Zeng17 seems to be saturated. However, we improved ATTC to 3.44 seconds with our AdaLEA. In the NIDB, our proposed configuration significantly improved from 92.5 [mAP] and 2.45 [ATTC] to 99.1 [mAP] and 4.81 [ATTC]. 
Although the values of both mAP and ATTC on the DAD tend to be lower, we believe that this is simply due to the data configuration. The first data configuration is positive data-size. The number of positives in the NIDB is 7.7 times bigger than the DAD (4,594@NIDB vs 596@DAD). These characteristics enable accidental scenes to be learned effectively. The second is an accident/incident scene configuration. While most of the videos in the NIDB include a simple near-miss incident between own vehicle and another risk-factor such as a cyclist or pedestrian, many videos in the DAD contain more complecated accidents between other risk-factors, without any danger to own vehicle.\par
Giving a careful consideration, a near-miss is appreciably related to a situation (e.g., cross road, rainy) so the model can partially solve this problem (see Table~\ref{table2}), however, to avoid accidents/incidents in advance, merely anticipating the presence of them is insufficient and more detailed information must be obtained in advance (e.g., what should be paid attention to).
Therefore, we further provide a more difficult task, risk-factor anticipation on the NIDB. Our method achieves 62.1 [mAP] and 3.65 [ATTC] which are +4.3 better with mAP and +0.70 earlier with ATTC than Chan~\textit{et al.}~\cite{ChanACCV2016}. We replace the loss function from EL to AdaLEA in Chan16+AdaLEA, after which we obtained better performance with mAP (+1.4) and ATTC (+0.28) in Table~\ref{table3}. Otherwise, in comparison to high mAP on risk anticipation, the mAP on risk-factor anticipation tend to be considerably lower, which implies that the model on risk anticipation focuses on the dangerous {\it situation}, almost not a target to be payed attention to as above mentioned.
  
  Moreover, in addition to AdaLEA and NIDB-pretrain, the QRNN on behalf of LSTM in Chan's method highly improves mAP and ATTC on both databases. This suggests that QRNN can focus on the direct relationship between frames (e.g., motion feature) and there is a possibiliy that QRNN is more suitable for analysis on consecutive sequential data, such as videos.

Figure~\ref{fig:vis} shows the visual comparison with EL, LEA and AdaLEA. Our proposed AdaLEA enabled a system to execute the earliest traffic accident anticipation. Our system anticipated when a car coming in the wrong direction appears at distance (the upper example) and when own vehicle is about to ignore the red signal (the lower example).

\section{Conclusion}

We presented a novel approach for traffic accident anticipation with our Adaptive Loss for Early Anticipation (AdaLEA) and self-annotated Near-miss Incident DataBase (NIDB) for anticipation. The AdaLEA allows a model to gradually learn an earlier anticipation as the training progresses. In our design, the loss adaptively assigns penalty weights  depending on how early a model can anticipate a traffic accident at each training epoch, inspired by Curriculum Learning. In the NIDB, we provide new task for risk-factor  anticipation. The NIDB also provides a better feature representation as NIDB-pretrain. 
With AdaLEA, NIDB-pretrain, and Quasi-RNN, our proposal achieved the best level of traffic accident anticipation performance in terms of mAP and ATTC. In a comparison with the conventional work, our proposal is +1.8@DAD, +6.6@NIDB better with mAP, and +0.43@DAD, +2.36@NIDB earlier with ATTC on risk anticipation. As for risk-factor anticipation in the NIDB, our proposed configuration was found to have improved from 57.8 [mAP] and 2.95 [ATTC] with the conventional work, to 62.1 (+4.3) [mAP] and 3.65 (+0.70) [ATTC].


{\small
\bibliographystyle{ieee}
\bibliography{nearmissanticipation}
}

\end{document}